\documentclass[11pt]{article}
\usepackage[T1]{fontenc}
\usepackage[utf8]{inputenc}
\usepackage[margin=1in]{geometry}
\usepackage{newtxtext}
\usepackage{amsmath}
\usepackage{newtxmath}
\usepackage{graphicx}
\usepackage{booktabs}
\usepackage{adjustbox}
\usepackage{authblk}
\usepackage{microtype}
\usepackage[font=small,labelfont=bf]{caption}
\usepackage[hidelinks]{hyperref}
\hypersetup{
  pdftitle={Europe's Climate Ambition Under Scrutiny: Evidence from Deep Learning Emission Projections},
  pdfauthor={Jacopo Ghirri, Carlos Rodriguez-Pardo, Lara Aleluia Reis, Massimo Tavoni}
}

\newcommand{\orgname}[1]{#1}
\newcommand{\orgaddress}[1]{#1}
\newcommand{\street}[1]{#1}
\newcommand{\city}[1]{#1}
\newcommand{\postcode}[1]{#1}
\newcommand{\state}[1]{#1}

\title{Europe’s Climate Ambition Under Scrutiny: Evidence from Deep Learning Emission Projections}

\author[1,2,3, *]{Jacopo Ghirri}

\author[1,2,3]{Carlos Rodriguez-Pardo}

\author[2,3,4]{Lara Aleluia Reis}

\author[1,2,3]{Massimo Tavoni}

\affil[1]{\orgname{Politecnico di Milano}, \orgaddress{\street{Piazza Leonardo da Vinci 32}, \city{Milan}, \postcode{20133}, \state{Italy}}}

\affil[2]{\orgname{CMCC Foundation - Euro-Mediterranean Center on Climate Change}, \orgaddress{\street{Via Marco Biagi 5}, \city{Lecce}, \postcode{73100}, \state{Italy}}}

\affil[3]{\orgname{RFF-CMCC European Institute on Economics and the Environment}, \orgaddress{\street{Via Bergognone 34}, \city{Milan}, \postcode{20144}, \state{Italy}}}

\affil[4]{\orgname{CENSE—Center for Environmental and Sustainability Research \& CHANGE—Global Change and Sustainability Institute, NOVA School of Science and Technology, NOVA University Lisbon}, \orgaddress{\city{Caparica}, \postcode{2829-516}, \state{Portugal}}}

\affil[*]{Corresponding Author. Email: jacopo.ghirri@polimi.it}

\date{}

\begin{document}

\maketitle
\abstract{The European Union has committed to reducing greenhouse gas emissions 55\% below 1990 levels by 2030, but whether current trends are compatible with this ambition remains uncertain. We apply deep learning to high-resolution socioeconomic and sectoral data across EU27 member states till 2023 to project sectoral CO$_2$ trajectories under current trends, extrapolating observed sectoral momentum without assuming changes in the pace or effectiveness of the policy environment beyond what is already reflected in historical data. We project that EU27 emissions will exceed the 2030 target by 35\% (620 Mt CO$_2$ shortfall), with only a small minority of countries on trajectories consistent with the bloc's commitments. While the Power sector achieves target-consistent reductions driven by the renewable transition, Mobility shows minimal progress and accounts for over a third of total emissions by 2030, reflecting a structural inertia across member states rather than geographically concentrated lag. Our findings indicate that substantial additional intervention is required to close Europe's ambition-implementation gap, and call for establishing up-to-date energy information in Europe.}

\section{Introduction}

Accurately forecasting short-term greenhouse gas emissions is key to assessing climate progress and re-orienting public policy towards effective mitigation strategies. In the European Union (EU), understanding the trajectory of sectoral emissions is essential for assessing whether the EU remains on course toward its 55\% reduction objective in 2030 \cite{meinshausen_realization_2022, riahi_cost_2021}, but also for identifying where progress is lagging and where additional policy action is needed. CO$_2$ emissions are determined by a high-dimensional mix of economic activities, energy use and consumption patterns, mobility demand, technological change, and climate variability \cite{jin_carbon_2024}, factors that evolve heterogeneously across countries and sectors. Capturing these dynamics in a systematic, data-driven, and interpretable way remains a major challenge \cite{feng_evcbilnet_2025, miraftabzdeh_deep_2025}, rooted in the limitations of both structural and machine-learning approaches to emissions modeling.

Structural models such as Integrated Assessment Models (IAMs), macroeconomic frameworks, and energy system optimization models are designed to evaluate energy and climate policies under internally consistent assumptions about the functioning of the energy system \cite{riahi_shared_2017}, making them well-suited for scenario analysis but less adequate for accurate short-term forecasting. 
In practice, imperfect information, limited access to finance, institutional rigidities, and heterogeneous objectives can substantially delay or attenuate policy responses \cite{jewell_political_2020, jewell_feasibility_2023}, and the realized short-term impact of carbon pricing or command-and-control instruments frequently diverges from model-based projections \cite{bertram_feasibility_2024, brutschin_multidimensional_2021}. 
Beyond this, IAMs and energy system optimization frameworks typically assume rational actors, perfect policy implementation, and a degree of intertemporal foresight that is rarely observed in practice \cite{jewell_political_2020, jewell_feasibility_2023, brutschin_multidimensional_2021}: empirical trend extrapolation provides a complementary lens precisely because it makes no such assumptions, grounding projections in the behavioral and structural patterns that have actually governed emission dynamics.

The availability of rich, sectoral-scale datasets has expanded substantially in recent years. Indicators compiled by Eurostat, the IEA, FAO, and Copernicus now cover a broad range of economic, energy, land-use, and climate variables at annual resolution across European countries. These data provide an unprecedented opportunity to build models that move beyond stylized scenarios and instead reflect empirical sectoral patterns. Despite these developments, existing forecasting tools have been slow to integrate such heterogeneous signals, and most approaches still rely on simplified assumptions or coarse representations that mask sector-specific drivers, though recent contributions suggest this gap is beginning to close \cite{way_empirically_2022, wen_hybrid_2020, jin_carbon_2024, ye_industrial_2024}.

Existing approaches span a spectrum from interpretable white-box models to opaque black-box methods \cite{jin_carbon_2024, feng_evcbilnet_2025, miraftabzdeh_deep_2025}. On one side of the spectrum, white-box models span a range from standard econometric regressions \cite{deng_how_2024, yang_forecasting_2018, yu_impact_2018, yu_impact_2019} to structural macro--energy frameworks \cite{riahi_shared_2017}. Econometric approaches can estimate historical relationships between emissions and their drivers, but typically rely on low-dimensional specifications that aggregate across sectors or assume stable functional forms, limiting their ability to capture the heterogeneous, nonlinear dynamics that characterize real emission trajectories \cite{deng_how_2024, bertram_feasibility_2024, luxembourg_times-europe_2025, yang_forecasting_2018, yu_impact_2018, yu_impact_2019}. 
On the other extreme, black-box machine-learning models have achieved high predictive accuracy, but at the expense of usability for policy insight. 
Autoregressive neural networks or models regressing directly on GDP, energy demand, and past emissions can fit historical data well \cite{vyas_impact_2013, wu_application_2018, nassef_application_2023, mutascu_co2_2022}, yet their internal representations are not always directly mappable to physical or economic drivers, limiting their value for diagnosing why emissions change or for understanding which structural factors are associated with emission dynamics \cite{jin_carbon_2024}.

These approaches thus leave a specific and policy-relevant question unanswered: given observed sectoral dynamics, what emissions trajectory is Europe currently on if current trends persist?

Answering this question also requires confronting a persistent limitation shared by both modeling families: the treatment of uncertainty as an add-on rather than a core modeling principle \cite{abdar_review_2021, miraftabzdeh_deep_2025}.
While uncertainty quantification has recently begun to attract attention in the climate economics field \cite{mastrandrea_ipcc_2011, intergovernmental_panel_on_climate_change_ipcc_climate_2023}, most data-driven forecasts remain deterministic, and most scenario-based models address uncertainty through discrete scenario comparison rather than probabilistic characterization, leaving statistical variability and structural ambiguity poorly understood and accounted for \cite{pao_forecasting_2012, li_can_2018, yang_short-term_2020, malik_forecasting_2020}. This tendency toward deterministic projection often leaves the range of plausible outcomes undercharacterized \cite{oreskes_climate_2010}. In scenario-based assessments, this limitation is compounded by the fact that projected pathways also depend on normative assumptions about policy ambition, technology deployment, and feasibility.

Emissions trajectories are determined by shocks, structural transitions, and policy changes that are deployed unevenly across time, space, and sectors. Models not accounting for such a structural and deep uncertainty risk overstating confidence in their predictions \cite{kendall_what_2017}, and are more prone to overfitting and fragile extrapolation. For climate policy, where decisions must remain robust under a wide range of plausible futures and where the cost of underestimating uncertainty can be severe, characterizing the spread of possible outcomes is as important as predicting central trends. 

Here, rather than specifying assumptions about future policy effectiveness, technology adoption rates, or demand trajectories, we track high-frequency sectoral indicators to assess whether the current policy portfolio is translating into measurable emission reductions consistent with stated goals. The resulting projections are best understood as a current-trends counterfactual, the trajectory Europe is on absent additional intervention, which complements rather than competes with scenario-based assessments of what policies could deliver.
\section{Framework Overview}

We forecast sectoral CO$_2$ emissions across EU27 member states from high-dimensional observational data spanning multiple indicators across energy production and consumption, fuel prices, electric vehicle sales, renewable capacity, land use, and international trade, among others (full list in Table \ref{tab:DataSources}). CO$_2$ is modeled as a proxy for total greenhouse gas emissions; non-CO$_2$ gases such as methane and nitrous oxide are excluded as their sectoral dynamics are governed by distinct and more complex processes that fall outside the scope of this study.
Given the high dimensionality of these inputs, we compress them into a structured latent representation using a variational autoencoder, which captures the core transition dynamics of each country-year while preserving natural variability. This latent representation, together with context variables such as GDP, population, and climate conditions for which reliable near-term projections exist, is then used to forecast emissions autoregressively through 2030. 
Uncertainty is propagated explicitly throughout, from the probabilistic latent encoding through to the final sectoral emission estimates.

We model CO$_2$ emissions as a function of high-dimensional, heterogeneous socioeconomic and sectoral indicators observed across EU27 member states. 
These indicators span different thematic domains: energy consumption and prices, electricity generation and trade, transportation of goods and people, economic indicators, land use and agriculture, and climate. The full set of variables, together with their sources, is reported in Table \ref{tab:DataSources}. All indicators are observed annually from 2010 through 2023; 2024 sectoral CO$_2$ emissions are available from Eurostat and used as an anchor for the projection chain, but the broader indicator panel for that year is not yet sufficiently complete to drive the encoder directly, a point we return to in Section \ref{sec_proj_procedure}.

Formally, for each country $i$ and year $t$, we observe a vector of transition indicators $x(t, i) \in \mathbb{R}^d$, alongside context variables $c(t, i)$. Our target is a vector of sector-wise CO$_2$ emissions $y(t, i) \in \mathbb{R}^s$ across $s=6$ sectors.

We identify three core challenges. First, $d$ is large relative to the number of country-year observations, and many indicators co-move in complex, nonlinear ways. Second, we aim to project emissions forward to 2030 without conditioning on assumed policy outcomes or structural transitions. We want to extrapolate empirical momentum, not simulate the outcomes of hypothetical policy interventions. Third, both emissions trajectories and transition indicators present noise and uncertainties, and a framework not accounting for these issues risks overfitting and failing to capture the true underlying mechanism guiding decarbonization.

\begin{figure}
    \centering
    \includegraphics[width=\linewidth]{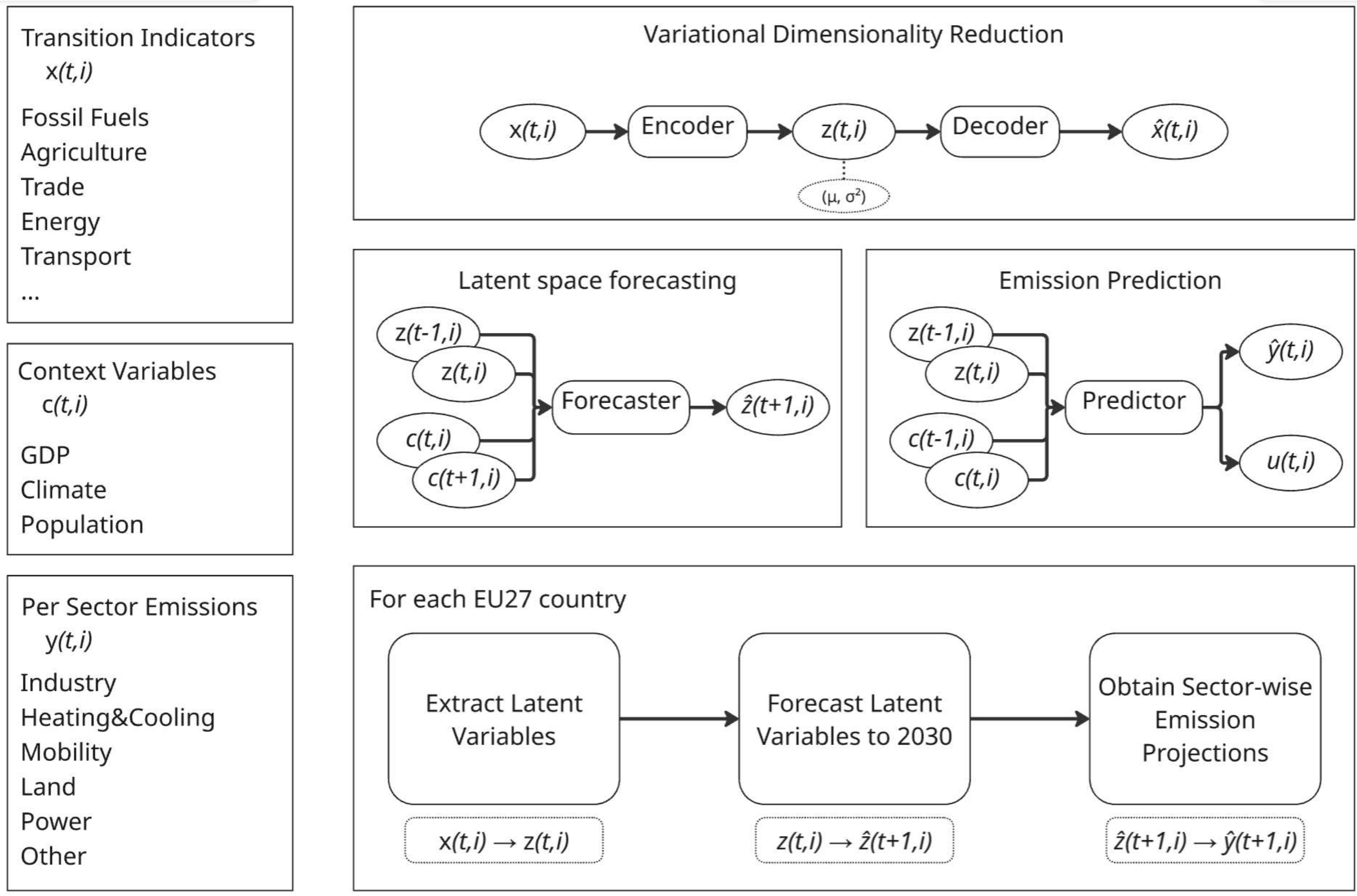}
    \caption{Overview of the modeling framework. The framework consists of three components: a VAE that maps high-dimensional transition indicators $x(t,i)$ to a probabilistic latent representation $z(t,i)$; a latent-space forecasting model that projects these representations forward using context variables $c(t,i)$; and an emission predictor that maps the projected states and context variables to sector-wise emission estimates $\hat{y}(t,i)$ and uncertainty outputs $u(t,i)$. Here $t$ denotes the year and $i$ the EU27 member state. For each country, observed indicators are encoded, iteratively projected to 2030, and decoded into sectoral CO$_2$ emission projections}
    \label{fig_structure}
\end{figure}

Our framework chains three components sequentially for each EU27 country, as reported in Figure \ref{fig_structure}.
A Variational Autoencoder (VAE) \cite{kingma_auto-encoding_2022} compresses each country-year observation $x(t,i)$ into a low dimensionality latent space $z(t,i)\sim\mathcal{N}(\mu, \sigma^2)$, acting as an internal state representation capturing both the central structure of the data and its natural variability. 
A latent forecaster then iterates $z(t,i)$ forward autoregressively from the last observed year through 2030, conditioned on context variables $c(t,i)$, for which there are reliable near-term projections. 
An emission predictor finally maps the projected latent sequence and context variable to sectoral emission estimates $\hat{y}(t,i)$, alongside a learned uncertainty vector $u(t,i)$ that penalizes overconfident extrapolation in sectors where emissions are structurally volatile. 
Full architectural and training details are given in the Methods section \ref{sec_method_models}.

\section{Results}
\paragraph*{A note on reported uncertainty}
All numerical projections are reported as point estimates accompanied by 90\% model consistency bands (m.c.b.), shown in brackets as [m.c.b. \textit{lower -- upper}].
These bands are derived from 10,000 Monte Carlo samples of the encoder's latent distribution, post-hoc calibrated to achieve 90\% coverage on historical residuals (see Methods Section \ref{sec_unce_calibration}). They reflect uncertainty in the mapping from observed indicators to emissions given the expected latent trajectory, and are not predictive distributions over future outcomes. In particular, they do not capture uncertainty in the latent forecasting step, future structural shocks, or policy discontinuities. They are most informative as comparative measures of relative predictability across sectors and countries, rather than as confidence bounds on realized 2030 emissions.

\subsection{Falling short of targets}
We project that, if current trends persist, EU27 CO$_2$ emissions are going to decline from 2.72 Gt in 2024 to 2.36 [m.c.b. 2.19 -- 2.53 Gt] Gt by 2030. A reduction of 13.2\% [m.c.b. -19.5\% -- -7\%], which is falling short of the 36\% reduction needed to abate 55\% of 1990 CO$_2$ emissions, constituting a gap of 622 Mt CO$_2$ [m.c.b. 454 -- 793 Mt] from the 1.74 Gt target. This target is derived by applying the 55\% reduction to 1990 CO$_2$ emissions alone, as a proxy for the FF55 commitment, which formally applies to all greenhouse gases and does not specify a CO$_2$-only reduction level. Our focus on CO$_2$ reflects both its dominance in greenhouse gas emissions and the greater regularity and tractability of its sectoral dynamics relative to non-CO$_2$ gases.

\begin{figure}
    \centering
    \adjustbox{width=1.3\linewidth, center}{
        \includegraphics{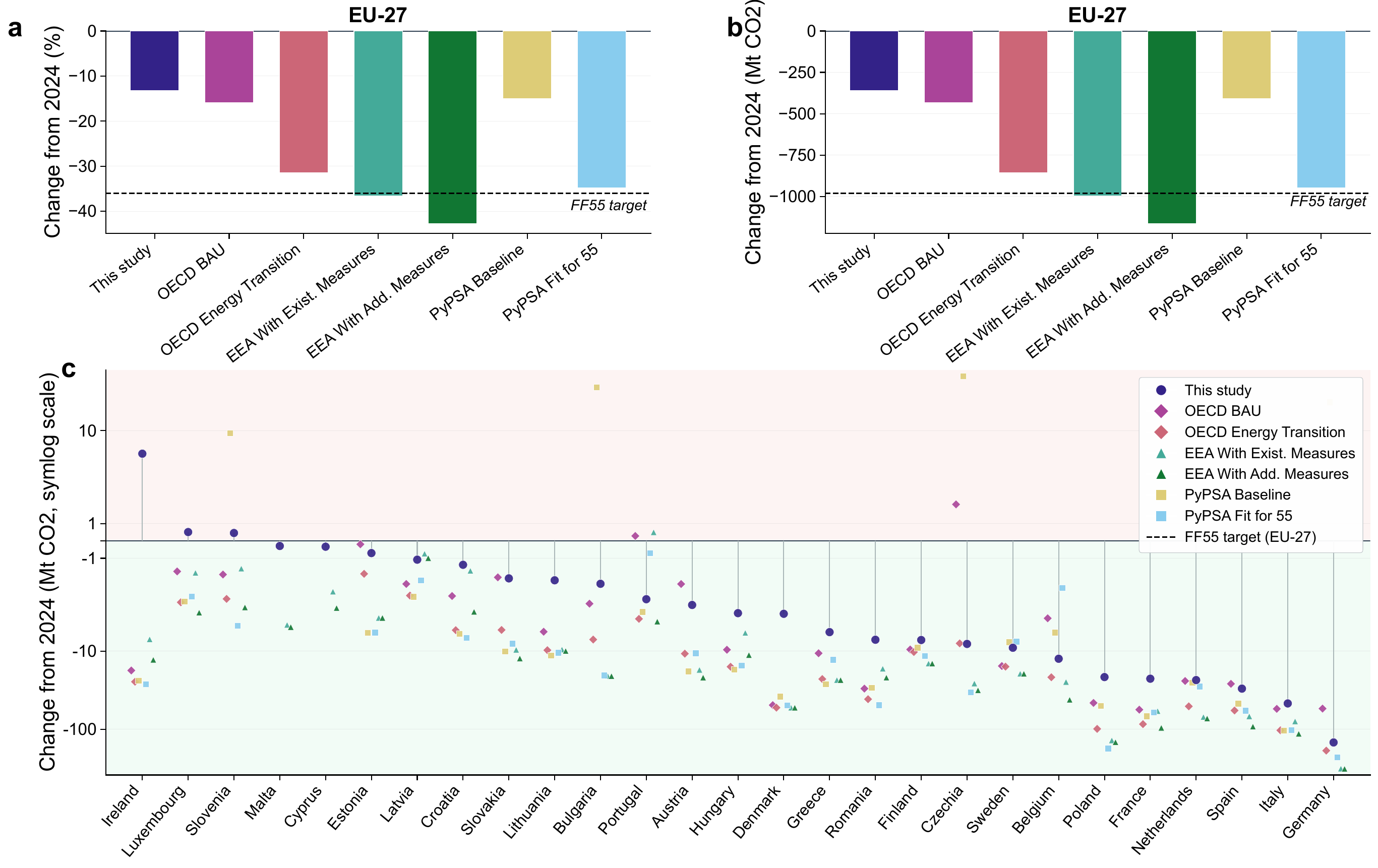}
    }
    \caption{Projected EU27, and all member states, CO$_2$ emission change from 2024 to 2030. Panels (a) and (b) show the projected percentage and absolute change in EU27 aggregate emissions relative to the 2024 baseline, respectively, for this study and six reference projections (OECD BAU, OECD Energy Transition, EEA With Existing Measures, EEA With Additional Measures, PyPSA Baseline, PyPSA Fit for 55). The dashed line indicates the change needed to meet a 55\% reduction from 1990 levels. Panel (c) shows the distribution of projected emission changes by member state on a symmetric logarithmic scale, with each source plotted separately per country. Countries are ordered by their projected change under this study.}
    \label{fig_emission_change}
\end{figure}

Figure \ref{fig_emission_change} contextualizes this finding against six alternative projection sources. Our estimate sits close to the OECD Business as Usual scenario (16\% emission reduction over the same period) and substantially above the most optimistic scenario, EEA With Additional Measures (43\% emission reduction).
This ordering is to be expected: third-party baselines, even conservative ones, incorporate forward-looking assumptions about policy effectiveness, technology adoption rates, and future demand trajectories that go beyond what is directly observable in historical data. Our framework makes no such assumptions in either direction, neither does it anticipate policy acceleration nor assumes policy rollback. Instead, it extrapolates the empirical patterns already present in the data at their observed pace. 
The alignment between our estimate and the OECD BAU, derived through an entirely different methodology, corroborates that current momentum places Europe well short of its 2030 commitments.

Panel c of Figure \ref{fig_emission_change} shows the distribution of projected emission changes from 2024 to 2030 across member states. Decarbonization is uneven: 10 countries show projected reductions exceeding 10\%, while the other 17 show slower decarbonization, or even increases.
The largest emitters, Germany, France, Italy, Poland, and Spain, collectively accounting for 65.5\% of 2024 emissions, achieve a collective decrease in emissions of 15\% [m.c.b. -24\% -- -6\%], with Germany showing the strongest decline among this group.

\subsection{Uneven progress}\label{sec_uneven}

\begin{figure}
    \centering
    \adjustbox{width=1.3\linewidth, center}{
        \includegraphics{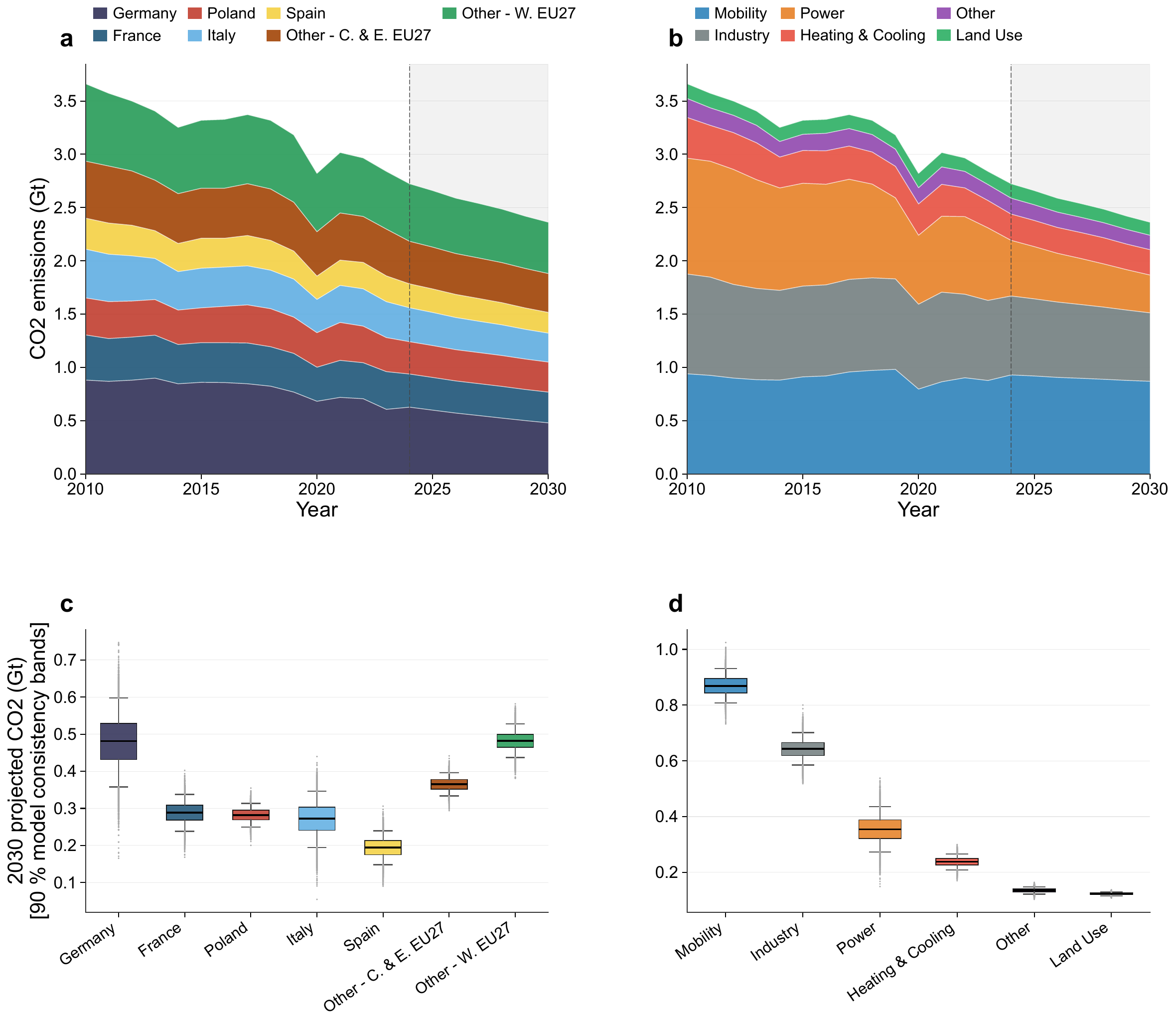}
    }
    \caption{EU27 CO$_2$ emission trajectories from 2010 to 2030, decomposed by country and sector, with projected 2030 uncertainty distributions. Panels (a) and (b) show historical (2010–2024) and projected (2025–2030) stacked area charts of total CO$_2$ emissions, decomposed by country group and sector, respectively. The dashed vertical line marks the boundary between the observed and projected periods. Panels (c) and (d) show box-and-whisker plots of the 2030 projected emissions across 10,000 Monte Carlo samples, by country group and sector. Boxes show the interquartile range and whiskers the 90\% model-consistency band, rescaled to match model residuals ditribution (see Methods Section \ref{sec_unce_calibration}).}
    \label{fig_emissions_decomposition}
\end{figure}

Figure \ref{fig_emissions_decomposition} decomposes the EU27 emission trajectories from 2010 to 2030 by country and sector, revealing substantial heterogeneity in decarbonization progress beneath the aggregate trend. 
Total EU27 emissions declined from 3.66 Gt in 2010 to 2.72 Gt in 2024, a 25.6\% reduction over the historical period. Our projections indicate a continued decline to 2.36 Gt by 2030, corresponding to a total reduction of 35.5\% [m.c.b. -40.1\% -- -30.8\%] from 2010 levels, but with markedly different contributions across sectors.

The Power sector stands out as the clearest decarbonization success, declining from 1.09 Gt in 2010 to a projected 0.35 Gt [m.c.b. 0.27 -- 0.44 Gt] in 2030 (67\% [m.c.b -74.9\% -- -59.9\%] reduction). This trajectory is broadly consistent with the scale of reduction required under Fit for 55 targets, and reflects the structural shift toward renewable generation that has accelerated across the EU over the past decade. In contrast, Mobility is projected to remain the largest emitting sector in 2030 at 0.87 Gt [m.c.b. 0.81 -- 0.93 Gt] (36.8\% of total [m.c.b. 35\% -- 38.7\%]), having declined only 7.5\% from 2010 [m.c.b. -14\% -- -0.8\%], the slowest trajectory of any sector. Industry shows a significant, but slower, decarbonization (31.4\% from 2010 [m.c.b. -37.5\% -- -25.1\%]), while Heating \& Cooling, Land Use, and Other sectors contribute smaller shares (21\% aggregate share of 2030 projected emissions [m.c.b. 19.7\% -- 22.4\%]), totaling 0.5 Gt in 2030 [m.c.b. 0.45 -- 0.54 Gt]. Notably, the COVID-19 pandemic produced a visible dip in emissions around 2020, particularly in Mobility, with post-pandemic recovery evident before the projected decline resumes.

\begin{figure}
    \centering
    \includegraphics[width=\linewidth]{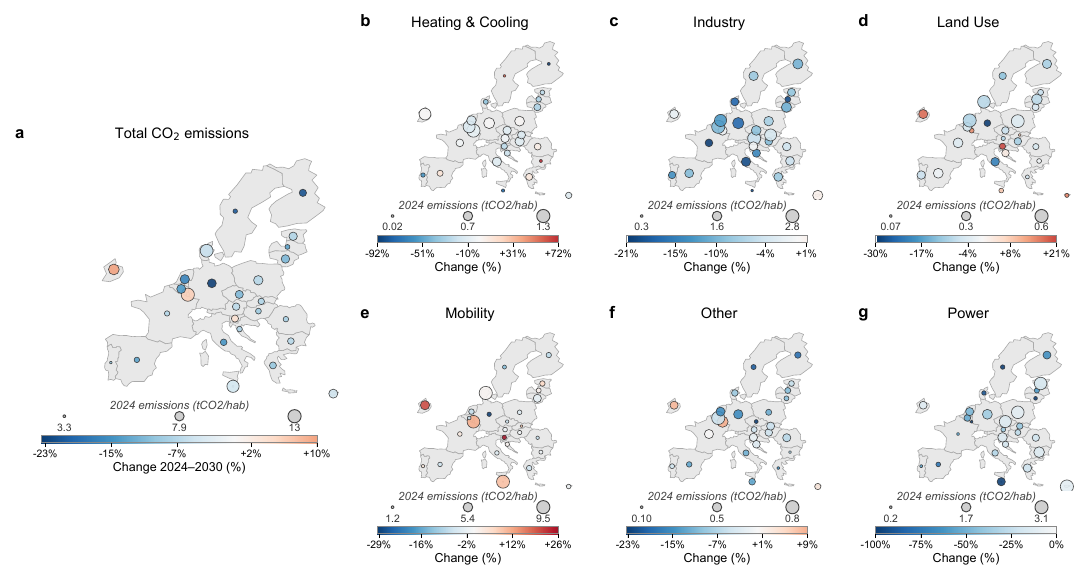}
    \caption{Projected total and sectoral CO$_2$ emission changes from 2024 to 2030 across EU27 member states. Panel a shows the aggregate change across all sectors. Panels b to g show changes for Heating \& Cooling, Industry, Land Use, Mobility, Other, and Power, respectively. Circle color encodes the direction and magnitude of change (blue: decreasing; red: increasing), with each panel using an independent color scale to reflect sector-specific ranges. Circle size encodes 2024 per capita emissions, with reference values shown below each panel.}
    \label{fig_sector_maps}
\end{figure}

Figure \ref{fig_sector_maps} confirms that this sectoral pattern holds broadly across member states, but with considerable heterogeneity. Power decarbonizes consistently: 11 of 27 countries show projected reductions exceeding 50\%, reflecting the structural clarity of the ongoing renewable transition. The picture for Mobility is starkly different, as 11 countries show projected increases, with no clear regional pattern, suggesting that the resistance to transport decarbonization is structural rather than geographically concentrated. Germany and Italy, the two largest contributors to the block's emissions, emerge as broad-based decarbonizers, achieving reductions across all six sectors. The Heating \& Cooling sector displays the highest cross-country variance, ranging from 92\% reduction to 72\% increase. These extreme swings are however concentrated in countries that already exhibit exceptionally low per capita emissions in this sector, such as Sweden, where the small absolute baseline amplifies percentage changes. Land Use similarly shows high cross-country variance. A full country-by-sector breakdown, comprehensive of model confidence, is reported in Figure \ref{fig_sectoral_heatmap} in the Supplementary Information.

\subsection{Model Interpretability}
\begin{figure}
    \centering
    \includegraphics[width=\linewidth]{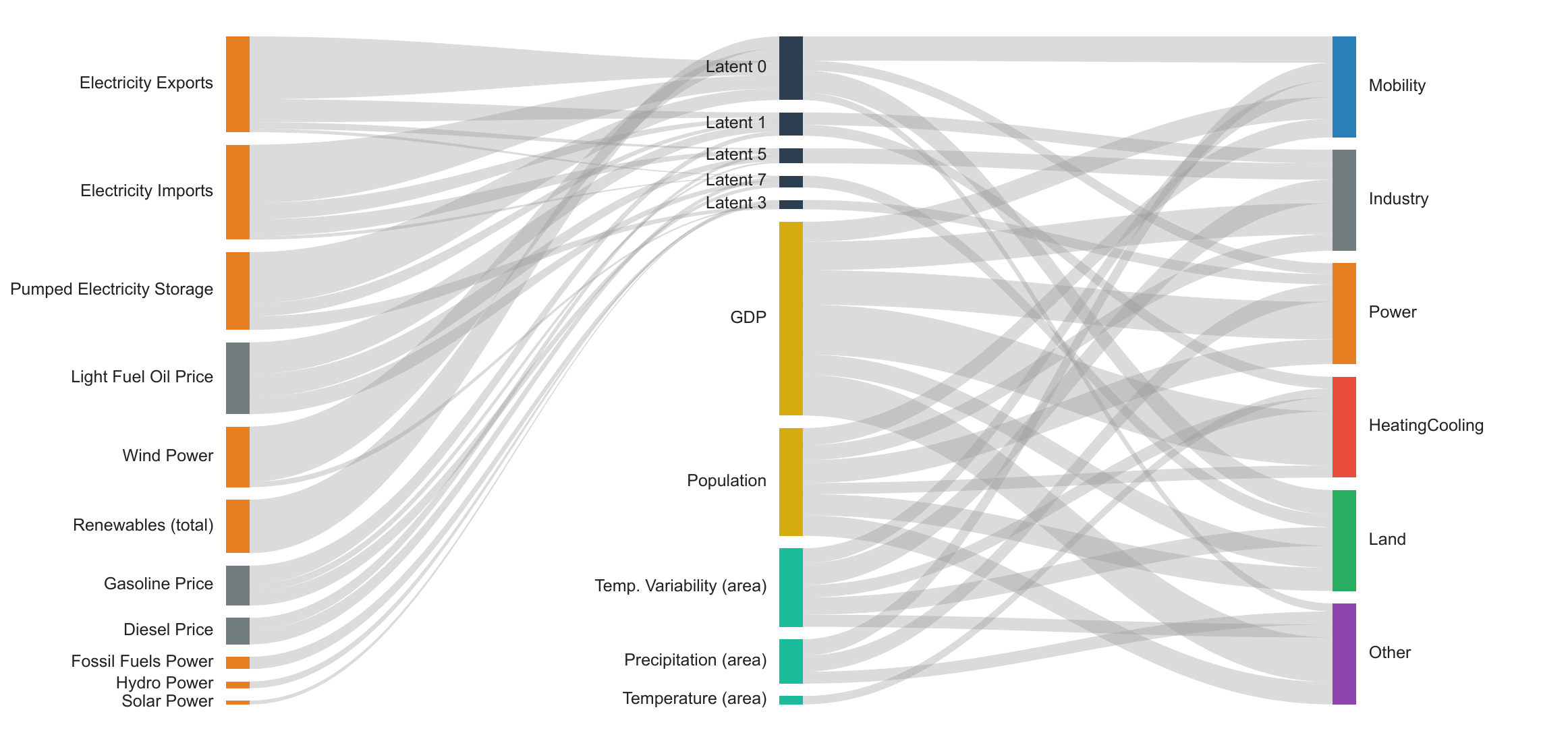}
    \caption{ Gradient-based attribution flows across the emission prediction pipeline. Input variables (left) connect through VAE latent dimensions (center) to sectoral emission outputs (right). Context variables enter the predictor directly, bypassing the encoder. For each emission output, the five most influential variables are shown; for each selected latent dimension, the five most influential input variables are shown. Node and edge widths are proportional to mean absolute gradient, averaged across all country-year observations in the training sample.}
    \label{fig_gradient_vis}
\end{figure}

To characterize which inputs drive the model's sectoral predictions, we compute gradient-based attributions across the encoder's dimensionality reduction ($\partial \mu_j / \partial x_i$) and the predictor's emission attribution ($\partial \hat{y}_s / \partial v_i$). Figure \ref{fig_gradient_vis} shows, for each output, the top five activations by absolute gradient, averaged across all country-year observations, and for each selected latent feature the top five input variables by gradient activation. Flows are proportional to the mean absolute gradient.

The attribution structure highlights a strong influence of context variables, particularly GDP and population, which enter the predictor directly and have broad activations across all sectors. 
Among the latent dimensions selected as most informative for emission predictions, there is a broad sensitivity to energy: from fuel pricing, to power generation and electricity trade. Moreover, we highlight two particular cases.
Latent 0 stands out as the most informative dimension, with the electricity balance (exports, imports, and pumped storage) emerging as its dominant input. This is consistent with energy balance serving as a compressed fingerprint of each country's structural energy position, with downstream consequences that extend across sectors. 
Latent 3, by contrast, is exclusively determined by the domestic generation mix, with no detectable contribution from trade or price variables, and connects strongly to power sector emissions, suggesting the VAE has isolated electricity generation as a dedicated and separable dimension of the latent space. 
Fossil fuel prices appear as the other major carrier of information, routing through the remaining active latent dimensions. Together, these results confirm the centrality of comprehensive energy policy, spanning generation mix, trade, and pricing, as the primary structural determinant of near-term emission trajectories across the EU.

\section{Conclusions}
The European Union's 2030 climate ambition faces a substantial implementation gap. Under current trends, we project that EU27 CO$_2$ emissions are going to reach 2.36 Gt by 2030, a 13\% reduction from 2024 levels that falls approximately 620 Mt CO$_2$ short of what is needed to achieve the Fit for 55 target. Our data-driven framework, which extrapolates empirical sectoral momentum without embedding policy assumptions, places Europe's trajectory closer to a business-as-usual baseline than to any scenario consistent with its stated goals. 
Decarbonization progress is highly uneven across member states, with only a small minority of countries on trajectories consistent with the block's commitments, indicating that the aggregate shortfall reflects a near-universal failure of current trends rather than the drag of a few lagging economies.

The sectoral picture is one of stark divergence. Power generation is on a structurally credible decarbonization path, projected to decline 67\% from 2010 levels by 2030, reflecting the accelerating renewable transition across the EU. Mobility, by contrast, remains deeply resistant to change: projected to account for 36.8\% of 2030 emissions with only a 7.5\% reduction since 2010, it represents the single largest obstacle to European climate targets. Crucially, this resistance is not geographically concentrated but broadly structural across member states, pointing to systemic rather than country-specific failures. Electric vehicles (EVs) offer the best and most affordable way to reduce emission from personal transportation. Our model is trained until 2023, thus incorporating the slowdown in EV sales observed by 2024. Currently, EV sales is ramping up again. It might be that these more recent trends, partly spurred by the energy crisis in the Middle East, will speed up the decarbonization of mobility, and that our estimates might be overstated, though it is worth reminding that freight emissions in Europe have been showing no sign of decline.

Our attribution analysis further suggests that energy system structure, generation mix, electricity trade balance, and fuel pricing, are the primary actionable determinants of near-term emission trajectories, pointing to these as the most plausible levers to shift current momentum.

These findings carry direct implications for European climate governance. The gap between current trajectories and stated commitments reflects the distance between the policy environment that has actually shaped emission dynamics and the more ambitious conditions assumed in optimistic scenarios. 
Closing this gap within the remaining timeframe will require interventions that go substantially beyond current policy portfolios, particularly in transport. Our results make clear that Europe's stated climate ambition is not yet matched by observable decarbonization momentum, especially in some crucial sectors such as transportation. Closing this gap demands urgent, targeted action rather than reliance on favorable assumptions about future policy effectiveness. To track the evolution of the energy system against its policy objectives, it is crucial to develop a database of energy and economic statistics that can enable near-real-time assessments, leveraging deep learning methods such as those used in this paper.

\subsection{Data Sources}

Table \ref{tab:DataSources} reports the data sources on which all models have been trained, and Table \ref{tab:EmissionSectors} reports the sectorial decomposition of emission sources.

\begin{table*}[htb!]
\centering
\footnotesize
\begin{tabular*}{\textwidth}{@{\extracolsep{\fill}}p{0.17\textwidth}p{0.62\textwidth}p{0.15\textwidth}@{}}
\toprule
\textbf{Category} & \textbf{Variables} & \textbf{Source} \\
\midrule
Emissions & CO$_2$ emissions by NACE activity and households (total and per capita) & Eurostat \\
Agriculture & Production of fruits, meats (total and poultry), vegetables, wheat & FAO \\
Energy Consumption & GJ per capita energy use (industry, transport, final consumption) & Eurostat \\
Economic Indicators & Energy taxes (€M), GDP (quarterly in €M), Population & Eurostat \\
Transportation & Modal split of freight (air, rail, road, sea), electric vehicle sales and stock, trains (diesel/electric) & Eurostat, IEA \\
Renewable Energy & Heat pumps (GWh), solar thermal collectors' surface & Eurostat \\
Land Use & Country area, coastal waters, cropland, forests & FAO \\
Electricity & Distribution losses, imports/exports, production (combustible, renewable, solar, wind, hydro) & IEA \\
Climate & Temperature (population/area weighted), rainfall, temperature variability & ERA5 (Copernicus) \\
Energy Prices & Gasoline, light fuel oil, automotive diesel prices (USD), EU ETS carbon prices & IEA, World Bank \\
International Trade & Export/import indices (chemicals, machinery, transport equipment, fuels, raw materials) & Eurostat \\
\bottomrule
\end{tabular*}
\caption{Input variables $x \in \mathbb{R}^d$ used in our study, and their sources. }
\label{tab:DataSources}
\end{table*}

\begin{table*}[!htb]
\centering
\footnotesize
\begin{tabular*}{\textwidth}{@{\extracolsep{\fill}}p{0.15\textwidth}p{0.81\textwidth}@{}}
\toprule
\textbf{Sector} & \textbf{Activity according to Eurostat definitions} \\
\midrule
HeatingCooling & Heating and cooling by households \\
\addlinespace
Industry & Manufacturing, mining, construction \\
\addlinespace
Land & Agriculture, forestry, fishing, water supply, waste management \\
\addlinespace
Mobility & Transportation, storage, trade, vehicle repair, transport by households \\
\addlinespace
Power & Electricity, gas, steam, air conditioning \\
\addlinespace
Other & Health, science, technical and professional services, finance, insurance, information, communication, public admin.,  arts, entertainment,  defense, administrative services, real estate, accommodation, food service, education, other services, and household activities \\
\bottomrule
\end{tabular*}
\caption{Target emission variables $\mathbf{y} \in \mathbb{R}^{s}$, gathered from Eurostat.}
\label{tab:EmissionSectors}
\end{table*}

\subsection{Framework and Modeling rationale}
We model sector-wise CO$_2$ emissions $\mathbf{y}(t,i) \in \mathbb{R}^s$ across $s=6$ sectors as a function of high-dimensional transition indicators $\mathbf{x}(t,i) \in \mathbb{R}^d$ and context variables $\mathbf{c}(t,i)$, observed annually for each EU27 member state $i$. The framework chains three learned components: an encoder $q_\phi$, a latent forecaster $f_\theta$, and an emission predictor $g_\psi$, trained sequentially on the historical panel and applied autoregressively to project sectoral trajectories through 2030.

The high dimensionality and nonlinear co-movement of $\mathbf{x}(t,i)$ motivate a learned latent representation rather than feature selection or standard dimensionality reduction. The encoder $q_\phi$ maps $\mathbf{x}(t,i)$ to a probabilistic latent state $\mathbf{z}(t,i) \sim \mathcal{N}(\boldsymbol{\mu}_\phi(\mathbf{x}), \boldsymbol{\sigma}^2_\phi(\mathbf{x}))$, encoding each observation as a distribution over latent space rather than a fixed point. Unlike deterministic alternatives such as PCA \cite{frs_liii_1901}, this probabilistic compression is robust to measurement noise and heterogeneous data quality, both pervasive features of cross-national indicator datasets, and preserves the cross-indicator covariance structure without requiring individual projections for each of the $d$ inputs.

Forecasting is performed in the latent space rather than the original indicator space, as the VAE's structured low-dimensional representation makes year-to-year transitions smoother and more regular than the raw indicators. The latent forecaster $f_\theta$ predicts:

$$\hat{\mathbf{z}}(t,i) = f_\theta\!\left(\mathbf{z}(t-1,i),\, \mathbf{z}(t-2,i),\, \mathbf{c}(t,i),\, \mathbf{c}(t-1,i)\right)$$

iterated from the last observed year through 2030. Context variables $\mathbf{c}(t,i)$, comprising GDP, population, and climate variables, enter the forecaster directly rather than through the encoder, as their near-term trajectories are comparatively tractable to project and their influence on emissions may operate through mechanisms not fully captured by the latent representation. The emission predictor $g_\psi$ then maps the projected latent sequence and context variables to sectoral outputs:

$$\hat{\mathbf{y}}(t,i) = g_\psi\!\left(\hat{\mathbf{z}}(t,i),\, \hat{\mathbf{z}}(t-1,i),\, \mathbf{c}(t,i),\, \mathbf{c}(t-1,i)\right)$$

alongside a learned uncertainty vector $\mathbf{u}(t,i) \in \mathbb{R}^s$, detailed in Section \ref{sec_methods_predictor}.

This framework does not embed policy targets, assume future technology adoption, or model behavioral responses to price signals. It learns emission dynamics as they have manifested historically and extrapolates those dynamics forward at their observed pace

\subsection{Model Architecture}\label{sec_method_models}
This section details our uncertainty-aware framework for modeling and interpreting macroeconomic drivers of sectoral carbon emissions. We begin by describing our variational autoencoder architecture that captures meaningful country-year representations, followed by our emission prediction model that leverages these latents, to provide both accurate predictions and uncertainty estimates, then we detail the model forecasting the latent features into the future.

\subsubsection{Variational Autoencoder}
The VAE maps transition indicators $\mathbf{x}(t,i)$ to a probabilistic latent state $\mathbf{z}(t,i) \sim \mathcal{N}(\boldsymbol{\mu}_\phi(\mathbf{x}), \boldsymbol{\sigma}^2_\phi(\mathbf{x}))$ \cite{kingma_auto-encoding_2022}.

\paragraph*{Encoder Network}
The encoder network maps input variables $x \in \mathbb{R}^d$ to parameters of a latent distribution $z \sim \mathcal{N}(\mu, \sigma^2)$ where $\mu, \log \sigma^2 \in \mathbb{R}^k$. We implement the encoder as a series of residual blocks connected by linear layers:

$$h_{\text{enc}}(x) = (\mu(x), \log \sigma^2(x))$$

\noindent Each residual block contains the following components:
\begin{itemize}
    \item A skip connection\cite{he_deep_2016} to facilitate gradient flow during training. These prove useful for stable training and fast convergence.
    \item Linear transformations with GELU\cite{hendrycks_gaussian_2016} activations, which empirically outperformed other non-linearities.
    \item Dropout\cite{srivastava_dropout_2014} regularization to prevent overfitting, applied both to neuron activations and to input features.
    \item No normalization layers, as we found empirically that they decreased model performance.
\end{itemize}

\paragraph*{Decoder Network}

The decoder mirrors the encoder's architecture but reverses the information flow, mapping samples from the latent distribution back to the original input space:

$$h_{\text{dec}}(z) = \hat{x}$$

We sample from the latent distribution using the reparameterization trick to enable backpropagation through the sampling process:

$$z = \mu(x) + \sigma(x) \odot \epsilon, \quad \epsilon \sim \mathcal{N}(0, I)$$

\paragraph*{VAE Loss Function}
We train the VAE using a weighted combination of reconstruction loss and KL divergence regularization:

$$\mathcal{L}_{\text{VAE}} = \lambda_{L1} \mathcal{L}_{\text{recon}} + \lambda_{KL} \mathcal{L}_{\text{KL}}$$

\noindent where $\lambda_{L1} = 0.95$ and $\lambda_{KL} = 0.05$ are weight hyperparameters. For the reconstruction loss $\mathcal{L}_{\text{recon}}$, we employ the L1 norm (Mean Absolute Error): $\mathcal{L}_{\text{recon}} = \|x - \hat{x}\|_1$. Empirically, we found that L1 yielded better convergence and more meaningful latent representations than MSE for our heterogeneous dataset.
The KL divergence term $\mathcal{L}_{\text{KL}}$ regularizes the latent distribution towards a standard Gaussian:

$$\mathcal{L}_{\text{KL}} = D_{\text{KL}}(\mathcal{N}(\mu(x), \sigma^2(x)) \| \mathcal{N}(0, I))$$

\paragraph*{Implementation Details}
The VAE was trained for 5000 epochs using AdamW\cite{loshchilov_decoupled_2017} with a learning rate of $5 \times 10^{-4}$, $\epsilon = 10^{-6}$, and a high weight decay of $0.1$. To ensure stable training, we applied gradient norm clipping to 1.0\cite{zhang_why_2019}. We use $0.33$ dropout in every layer and $0.4$ input dropout. We use 5 residual blocks, with 6 hidden layers each for the encoder, with a number of neurons linearly decreasing in each block, from the input dimension to the latent size. Hyperparameters were found following a Bayesian optimization.
Using CodeCarbon\cite{courty_mlco2codecarbon_2024}, we measure our electricity consumption: training the VAE uses 5.46 Wh of electricity, which amount to emitting 1.81 g CO$_2$eq at the carbon intensity of Italy's national electricity grid.

\subsubsection{Emissions Prediction Model}\label{sec_methods_predictor}
The emission predictor $g_\psi$ maps projected latent states and context variables to sectoral emission estimates, and is trained after the VAE on its fixed latent representations. We pair $g_\psi$ with an uncertainty-aware loss \cite{kendall_what_2017} that trains it to jointly estimate sectoral outputs and a per-sector confidence metric, penalizing large residuals while permitting higher uncertainty where emissions are intrinsically more variable or difficult to extrapolate.

\paragraph*{Temporal Context Integration}
To capture temporal dynamics in emissions patterns, our predictor incorporates both current and historical information. For each country-year data point, we sample two latent representations:
\begin{enumerate}
    \item The current year's latent state $z_t \sim \mathcal{N}(\mu(x_t), \sigma^2(x_t))$
    \item The previous year's latent state $z_{t-1} \sim \mathcal{N}(\mu(x_{t-1}), \sigma^2(x_{t-1}))$
\end{enumerate}

For cases where previous year data is unavailable (e.g., the first year in our dataset for each country), we use zero vectors as placeholders. Additionally, we include context variables from both the current and previous year explicitely, since these may impact emissions through mechanisms not fully captured by the latent representation. This approach allows the model to identify both absolute states and year-to-year changes that affect emissions.

The combined input to the predictor network is:
$$v = [z_t; z_{t-1}; c_t; c_{t-1}]$$
where $c_t$ and $c_{t-1}$ represent climate variables for the current and previous years, respectively, and $[;]$ denotes vector concatenation.

\paragraph*{Dual Output Architecture}
Our predictor returns two outputs:
\begin{itemize}
    \item Emissions predictions $\hat{y} \in \mathbb{R}^{2s}$, consisting ofper capita emissions across $s=6$ sectors
    \item Uncertainty estimates $\sigma_y \in \mathbb{R}^s_+$, representing the model's confidence in its predictions for each sector
\end{itemize}

\noindent This architecture follows a similar design to the VAE, employing residual blocks with GELU activations and dropout regularization. We use a network with wider hidden layers compared to the VAE, as experiments indicated this improved predictive performance.

\paragraph*{Uncertainty-Aware Loss Function}
We train the emissions predictor using an uncertainty-aware loss function\cite{kendall_what_2017} that balances prediction accuracy with uncertainty calibration. The loss function combines mean squared error with learned uncertainty, following the principled approach proposed in uncertainty quantification literature:

$$\mathcal{L} = \frac{1}{2} \exp(-\log \sigma^2_y) \cdot (y - \hat{y})^2 + \frac{1}{2} \log \sigma^2_y$$

This formulation has two key properties. First, it automatically balances the error term with the uncertainty regularization. When prediction errors are high, the model is incentivized to increase its uncertainty estimates. Conversely, $\log \sigma^2_y$ prevents the model from assigning arbitrarily high uncertainty to avoid the error penalty. By jointly optimizing for total and per capita emissions with shared uncertainty estimates, the model can develop a more robust representation of emission drivers while appropriately quantifying uncertainty.

\paragraph*{Implementation Details}
We trained the predictor for 5000 epochs using AdamW \cite{loshchilov_decoupled_2017} with a learning rate of $3 \times 10^{-5}$, $\epsilon = 10^{-6}$, and high weight decay of 0.22. As with the VAE, we applied gradient norm clipping to 1.0 to ensure stable training. We use dropout of $0.09$. We use 2 residual blocks with 2 hidden layers each, with 128 neurons each and SILU activation \cite{elfwing_sigmoid-weighted_2017}. We train all our models using PyTorch\cite{paszke_pytorch_2019}. Hyperparameters were found following a Bayesian optimization with WandB\cite{biewald_experiment_2020}.
Using CodeCarbon\cite{courty_mlco2codecarbon_2024}, we measure our electricity consumption: training the predictor uses 4.61 Wh of electricity, which amount to emitting 1.52 g CO$_2$eq at the carbon intensity of Italy's national electricity grid.

\subsubsection{Latent-Space Forecasting Model}
The latent forecaster $f_\theta$ learns the temporal evolution of the VAE latent space, enabling autoregressive projections that preserve the cross-indicator covariance structure encoded by $q_\phi$.

\paragraph*{Temporal Context Integration}

To model temporal dependencies, the forecasting network leverages information from multiple previous years. For each country-year couple, we construct an input vector that combines:
\begin{enumerate}
    \item A sample of the latent state of the previous year: $z_{t-1}=\mu(x_{t-1})$.
    \item A sample of the latent state two years prior: $z_{t-2}=\mu(x_{t-2})$.
    \item Climate and macroeconomic context variables for the current year $c_t$
    \item The same context variables from the previous year $c_{t-1}$
\end{enumerate}

The concatenated input of the forecast model is:
$$u = [z_{t-1}; z_{t-2}; c_t; c_{t-1}]$$

\paragraph*{Forecast Network Architecture}

The forecasting model $f_\theta$ maps $u_t$ to a prediction of the next latent representation:
$$\hat{z_t}=f_\theta(u_t)$$

We implement $f_\theta$ as a compact feed-forward network employing residual \cite{he_deep_2016} blocks, GELU \cite{hendrycks_gaussian_2016} activations, and dropout \cite{srivastava_dropout_2014} regularization—mirroring the architectural principles used in the emissions predictor. Residual connections facilitate stable training by allowing gradients to propagate through the forecast model effectively. The network predicts a vector in the latent space dimension, ensuring compatibility with both the VAE encoder and the emissions predictor.

\paragraph*{Learning Samples}

The predicted latent state $z^t$ aims to be a direct sample from the target latent distribution. We perform this by training using the uncertainty-aware loss function\cite{kendall_what_2017} balancing closeness with the target mean $\mu(x_t)$, and calibrating with the target variance $\sigma^2(x_t)$.

For each latent dimension, we compute the loss as:

$$\mathcal{L} = \frac{1}{2} \exp(-\log \sigma_y^2(x_t)) \cdot |\mu_y(x_t) - \hat{z}_{y,t}| + \frac{1}{2} \log \sigma^2(x_t)$$
This loss stabilizes training and encourages the forecaster to respect the probabilistic structure encoded by the VAE, rather than overfitting deterministic latent trajectories.

\paragraph*{Implementation Details}
We trained the forecaster for 5000 epochs using AdamW \cite{loshchilov_decoupled_2017} with a learning rate of $5 \times 10^{-4}$, $\epsilon = 10^{-6}$, and high weight decay of 0.09. As with the other models, we applied gradient norm clipping to 1.0 to ensure stable training. We use dropout of $0.15$. We use 3 residual blocks with 5 hidden layers each, with 128 neurons each and GELU activation \cite{hendrycks_gaussian_2016}. We train all our models using PyTorch\cite{paszke_pytorch_2019}. Hyperparameters were found following a Bayesian optimization with WandB\cite{biewald_experiment_2020}. 
Using CodeCarbon\cite{courty_mlco2codecarbon_2024}, we measure our electricity consumption: training the forecaster uses 29.86 Wh of electricity, which amount to emitting 9.87 g CO$_2$eq at the carbon intensity of Italy's national electricity grid.

\subsection{Projection Procedure}\label{sec_proj_procedure}
Emission projections from 2025 to 2030 are generated autoregressively for each EU27 member state.
The latent forecasting chain is initialized from the 2022 and 2023 observed indicator data, which constitute the last two years of the training set. For each projection year, the latent forecaster $f_\theta$ advances the latent state forward by one step, conditioned on projected context variables, and the emission predictor $g_\psi$ then maps the resulting latent state to per-sector emission change relative to the previous year.

The 2024 projection year requires special treatment. While observed sectoral CO$_2$ emissions are available for 2024 from Eurostat, the broader indicator dataset used to train and drive the encoder is not yet sufficiently complete to robustly condition the latent chain from that year forward. We therefore use the latent forecaster to obtain a 2024 latent representation, which serves as the initialisation point for the 2025--2030 projection chain. Emission deltas are accumulated from 2025 onward, anchored to the observed 2024 sectoral values. This ensures that projections are grounded in the most recent available emission data, despite the latent representation for that year being derived from the forecaster rather than directly encoded from indicators.

Uncertainty is introduced at the emission prediction step only: for each Monte Carlo sample, the encoder's latent distribution $z(t,i) \sim \mathcal{N}(\mu(x), \sigma^2(x))$ is resampled at inference time, while the latent forecasting chain is held fixed at the posterior mean $\mu(x)$. 
This captures uncertainty in the mapping from observed indicators to the latent state and onward to sectoral emissions, but does not encompass uncertainty in the latent dynamics themselves. 
The resulting intervals therefore reflect uncertainty in the emission mapping given the expected development of each EU27 member state, rather than uncertainty over the trajectory itself.

\subsection{Uncertainty Interval Calibration}\label{sec_unce_calibration}
The framework produces two distinct uncertainty representations. 
The learned output $\mathbf{u}(t,i)$ of the emission predictor is a model-internal indicator of relative sectoral predictability. Separately, the Monte Carlo confidence intervals reported in Figure \ref{fig_emissions_decomposition} are obtained by repeatedly resampling the encoder's latent distribution when predicting emissions, as described below.

The Monte Carlo projections sample the encoder's latent distribution $z_t \sim \mathcal{N}(\mu(x_t), \sigma^2(x_t))$ repeatedly at the emission prediction step, with the latent forecasting chain fixed at 
the posterior mean $\mu(x_t)$, generating a spread over emission outcomes that reflects uncertainty in the mapping from observed indicators to the latent state and onward to sectoral emissions. This source of uncertainty is partial: it does not encompass uncertainty in the latent forecasting step, the predictor's learned sector-level confidence, or structural sources such as future shocks or measurement error in the input indicators. The intervals are therefore not predictive distributions over future outcomes but model-consistency bands, and are most informative for comparing relative predictability across sectors and countries.

Within this scope, the raw encoder spread need not be consistent with the model's empirical prediction errors. Following \cite{kuleshov_accurate_2018}, we apply a post-hoc rescaling to ensure that the reported bands achieve 90\% coverage on the historical period 2010–2023. For each country-sector cell $(i, s)$ we draw $N = 500$ emission samples per historical observation by re-sampling $z_t$ and passing each sample through the predictor, yielding an empirical mean $\hat{y}_{i,s,t}$ and standard deviation $\hat{\sigma}_{i,s,t}$. The standardised residual $r_{i,s,t} = (y_{i,s,t} - \hat{y}_{i,s,t})\,/\,\hat{\sigma}_{i,s,t}$ normalises prediction errors by the model's estimated spread, allowing direct comparison against a standard normal distribution. We estimate the temperature scalar:
$$T_{i,s} = \frac{quantile_{0.90}(|r_{i,s,t}|)}{1.645}$$

and rescale each MC sample as $y_{\text{cal},k} = \bar{y} + T_{i,s}(y_k - \bar{y})$, leaving the mean unchanged while adjusting the spread to match the empirical residual distribution. $T_{i,s} > 1$ indicates that the raw encoder spread underestimates historical errors; $T_{i,s} < 1$ indicates over-dispersion. Aggregate intervals sum the calibrated sample vectors across sectors before taking percentiles, preserving the joint distribution rather than summing marginal quantiles independently.

\subsection{Limitations}

Several limitations of the present framework should be kept in mind when interpreting our projections.

First, our framework models CO$_2$ emissions only, excluding non-CO$_2$ gases such as methane and nitrous oxide whose sectoral dynamics are more erratic and governed by distinct processes. Comparisons against the FF55 target, which formally applies to all greenhouse gases, should therefore be interpreted as indicative rather than exact.

Second, the indicator panel underlying our projections is observed only through 2023, with 2024 reconstructed via the forecaster rather than directly encoded (Section \ref{sec_proj_procedure}). This lag means the model cannot reflect the most recent shifts in policy, prices, or technology adoption. As new annual data become available, the framework can be re-estimated to narrow this gap, but, at any point in time, the model's most recent panel will lag the present by the publication delay of its training data sources.

Third, by design, the framework extrapolates empirical momentum without conditioning on future policy interventions, behavioral responses, or structural shocks. It will therefore systematically miss two classes of discontinuities: exogenous shocks of the kind documented in the rolling-origin evaluation, such as the COVID-19 pandemic, and abrupt changes in the pace or ambition of national climate policy that have no precedent in the historical record.
The resulting projections are best interpreted as a current-trends counterfactual rather than a forecast of realized outcomes.

Fourth, while we make a substantial effort to account for uncertainty, producing both learned sector-level confidence scores and calibrated Monte Carlo consistency bands, we acknowledge that these fall short of formal prediction intervals in the classical statistical sense. 
Model confidence scores are best interpreted as internal comparative measures of relative sectoral predictability, and are not safely translatable into probabilistic statements. The consistency bands, while calibrated against historical model errors, cannot be confidently interpreted as covering the true range of future outcomes, and should not be treated as predictive distributions over the projection horizon.

Finally, while the gradient-based attribution and sensitivity analyses provide high-level insight into the model's internal structure, they do not constitute a causal inference analysis. 
The associations identified between transition indicators and sectoral emissions reflect patterns learned from historical co-movement rather than structural causal relationships, and, while useful for validating the model's mechanisms and gathering high-level insights into the drivers of emission dynamics, should not be interpreted as explicit policy levers or quantitative estimates of the effect of specific interventions on emission trajectories.

\section*{Acknowledgments}

J. Ghirri, C. Rodriguez-Pardo, and M. Tavoni acknowledge funding from the European Union European Research Council (ERC), Grant project No 101044703 (EUNICE).

\bibliographystyle{plain}
\bibliography{CO2_forecast_EU}

\section*{Supplementary}

\subsection*{Latent representation inspection}
\begin{figure}
    \centering
    \includegraphics[width=\linewidth]{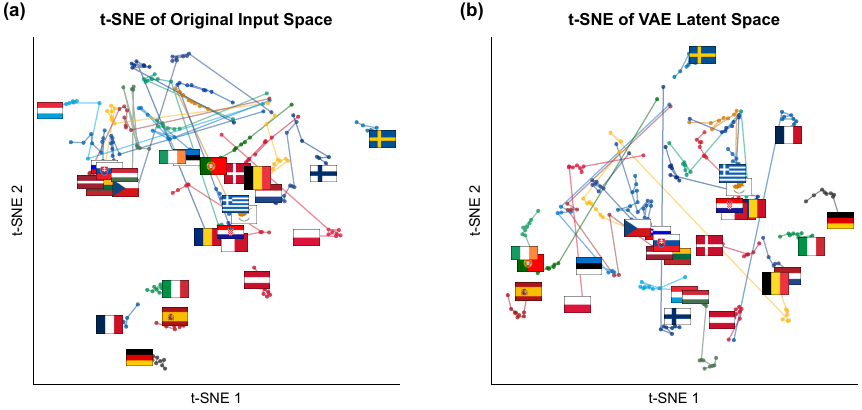}
    \caption{t-SNE projections of (a) the original high-dimensional input space and (b) the VAE latent space across all EU27 countries and years. Each point represents a country-year observation; connected points trace the temporal trajectory of each country. The latent space shows markedly improved separation between countries compared to the dense clustering observed in the original input space, indicating that the VAE learns a more structured and disentangled representation of emission-relevant characteristics.}
    \label{fig_tsne}
\end{figure}

\begin{figure}
    \centering
    \includegraphics[width=\linewidth]{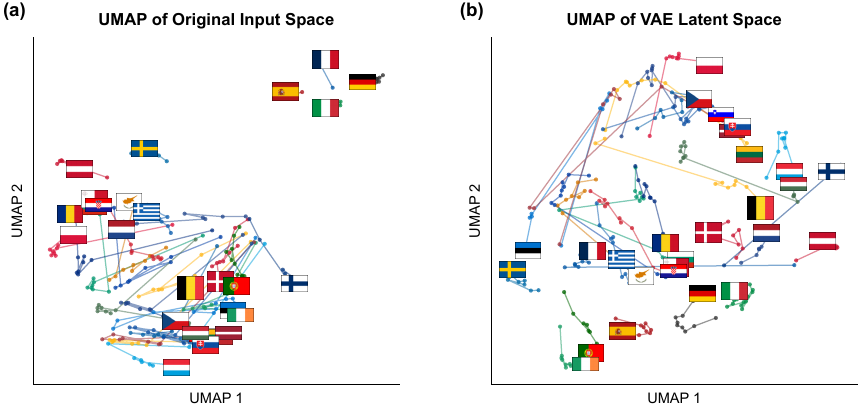}
    \caption{UMAP projections of (a) the original high-dimensional input space and (b) the VAE latent space across all EU27 countries and years. Each point represents a country-year observation; connected points trace the temporal trajectory of each country. Compared to the original space, the latent representation exhibits greater country separation and more continuous temporal trajectories, reflecting the VAE's ability to organise countries along meaningful structural dimensions while preserving global neighbourhood relationships.}
    \label{fig_umap}
\end{figure}

To assess the quality of the VAE latent representation, Figures \ref{fig_tsne} and \ref{fig_umap} compare the structure of the original input space and the learned latent space using t-SNE \cite{van_der_maaten_visualizing_2008} and UMAP \cite{mcinnes_umap_2018} dimensionality reduction respectively.

In the original high dimensional input space, both visualization reveal a characteristic pattern: the largest economies separate as clear outliers, driven primarily by scale differences, while most other member states collapse into a dense, poorly differentiated central cluster.
The VAE latent space substantially improves this structure. Country-level trajectories become more spatially separated and temporally coherent, with nations that share similar decarbonisation profiles, such as the Nordic countries or the coal-dependent Central European states, clustering more naturally. This suggests that the VAE learns to organise countries along transition-relevant structural dimensions rather than being dominated by raw economic scale.
Nevertheless, the latent trajectories are not perfectly smooth: abrupt jumps remain visible for several countries, reflecting genuine structural discontinuities in the underlying data, most notably the COVID-19 demand shock in 2020 and the energy price disruptions following the onset of the war in Ukraine in 2022, both of which produced sharp, simultaneous shifts across multiple input indicators. 
These discontinuities are not artifacts of the compression but rather faithful reflections of real regime disruptions that the VAE appropriately preserves rather than smooths away. Overall, both t-SNE and UMAP confirm that the  low dimensional latent space retains meaningful neighborhood structure while achieving a more disentangled and interpretable organization of the country-year observations.


\subsection*{Model Performance}
Performance metrics for all three model components are reported in Table \ref{tab:training_metrics}, computed on held-out validation sets corresponding to a 15\% random split of the training data, using the same split as model selection during training. All predictor metrics are reported in the z-score normalized emission space used during training (normalized kg CO$_2$ per capita); the latent forecaster metrics are in the normalized latent space. These units are not directly comparable to physical CO$_2$ quantities but provide a consistent basis for assessing relative fit across sectors and model components.

\begin{table}[htbp]
\centering
\caption{Model performance on held-out validation sets (15\% random split, seed 0). 
Predictor metrics are reported in the z-score normalised emission space used during 
training (normalised kg CO$_2$ per capita). Latent forecaster metrics are in the 
normalised latent space.}
\label{tab:training_metrics}

\vspace{0.5em}

\textbf{Panel A: Variational Autoencoder}

\vspace{0.3em}
\begin{tabular}{lcc}
\toprule
\textbf{Metric} & \textbf{Train} & \textbf{Validation} \\
\midrule
Reconstruction loss (L1)        & 0.124 & 0.179 \\
KL divergence                   & 0.819 & 0.818 \\
Mean latent mean                & 0.006 & 0.013 \\
Mean latent std                 & 0.580 & 0.578 \\
\bottomrule
\end{tabular}

\vspace{1.2em}

\textbf{Panel B: Emission Predictor (validation split)}

\vspace{0.3em}
\begin{tabular}{lccccccc}
\toprule
\textbf{Metric} & \textbf{H\&C} & \textbf{Industry} & \textbf{Land} & \textbf{Mobility} & \textbf{Other} & \textbf{Power} & \textbf{Aggregate} \\
\midrule
MAE  & 0.070 & 0.095 & 0.062 & 0.083 & 0.103 & 0.086 & \textbf{0.083} \\
RMSE & 0.110 & 0.128 & 0.092 & 0.139 & 0.169 & 0.128 & \textbf{0.130} \\
MSE  & 0.012 & 0.016 & 0.009 & 0.019 & 0.029 & 0.016 & \textbf{0.017} \\
R$^2$ & 0.985 & 0.972 & 0.991 & 0.988 & 0.969 & 0.964 & \textbf{0.981} \\
\bottomrule
\end{tabular}

\vspace{1.2em}

\textbf{Panel C: Latent Forecaster}

\vspace{0.3em}
\begin{tabular}{lcc}
\toprule
\textbf{Metric} & \textbf{Train} & \textbf{Validation} \\
\midrule
MSE  & 0.0062 & 0.0065 \\
RMSE & 0.0784 & 0.0805 \\
MAE  & 0.0470 & 0.0486 \\
\bottomrule
\end{tabular}

\end{table}

The VAE achieves stable reconstruction with near-identical KL divergence across train and validation splits (0.819 vs 0.818), indicating that regularization toward the standard Gaussian prior is consistent and does not depend on the specific observations seen during training. The small gap between training and validation reconstruction loss (0.124 vs 0.179) is consistent with expectations for a heterogeneous panel dataset where some country-year combinations are inherently harder to reconstruct.

The emission predictor achieves an aggregate R$^2$ of 0.981 on the validation set, with per-sector values ranging from 0.964 (Power) to 0.991 (Land). Notably, validation metrics are slightly better than training metrics across all sectors (aggregate R$^2$ of 0.969 on train vs 0.981 on val), which we attribute to the heavy regularization applied during training. High dropout rates create a harder effective training problem than the validation evaluation, which is conducted with dropout disabled. This pattern provides reassurance against overfitting. The relatively lower R$^2$ for Power and Other on the validation set is consistent with the higher uncertainty estimates learned by the predictor for these sectors, as reported in Section 4.2 of the main text.

The latent forecaster shows minimal generalization gap, with train and validation MSE of 0.0062 and 0.0065 respectively. This near-identical performance across splits reflects the smoothness and regularity of latent trajectories learned by the VAE: because the encoder compresses heterogeneous inputs into a low-dimensional structured space, year-to-year transitions in this space are predictable and consistent across countries. We note that the random validation split used here does not constitute a temporal or spatial out-of-sample test for the forecaster.

\subsection*{Rolling-Origin evaluation of the Forecasting Procedure}
To assess whether the pipeline generalizes to genuinely unseen time periods rather than interpolating within the training window, we conduct a rolling-origin evaluation in which all three model components, VAE, emission predictor, and latent forecaster, are retrained from scratch on data up to a cutoff year $T$ and evaluated on autoregressive projections for years $T+1$ through $T+3$. We repeat this procedure for $T \in \{2018, 2019, 2020, 2021\}$, generating emission projections for each held-out window and comparing against observed emissions from the full dataset. All metrics are reported in the z-score normalised emission space used during training.

A critical interpretive note is required before reading these results. This evaluation compares model projections against observed emissions in years where real policies were implemented, behavioral responses occurred, and structural shocks materialized, notably the COVID-19 demand collapse in 2020 and the energy price disruption following the onset of the war in Ukraine in 2022. By construction, our framework does not attempt to predict these events: it extrapolates empirical momentum assuming no further changes beyond natural evolution, producing a counterfactual trajectory rather than a forecast of what actually happened. Rolling-origin errors therefore reflect two distinct sources of divergence: genuine model error in extrapolating latent dynamics, and the gap between the counterfactual trajectory and a reality in which interventions and shocks did occur. These two components cannot be cleanly separated, but their joint presence means that the RMSE figures reported here should be interpreted as upper bounds on the model's extrapolation error under stable conditions, rather than as estimates of forecast accuracy in the conventional sense.

Table \ref{tab:rolling_origin} reports aggregate RMSE by training cutoff and forecast horizon. Mean RMSE increases monotonically with horizon, from 0.227 at one year ahead to 0.326 at three years ahead, consistent with the expected accumulation of errors in an autoregressive system. The absolute magnitude of this degradation is modest: mean RMSE at horizon +3 is approximately 40\% larger than at horizon +1, suggesting that the latent space learned by the VAE evolves sufficiently smoothly that short-term extrapolation remains coherent across the three-year window relevant for our 2030 projections.

\begin{table}[htbp]
\centering
\caption{Rolling-origin evaluation: aggregate RMSE (normalised emission 
space) for autoregressive projections of 1, 2, and 3 years ahead. Each row corresponds to models retrained from scratch on data up to the indicated cutoff year.}
\label{tab:rolling_origin}
\begin{tabular}{lccc}
\toprule
\textbf{Training cutoff} & \textbf{Horizon +1} & \textbf{Horizon +2}     & \textbf{Horizon +3} \\
\midrule
2018    &   0.282   &   0.416   &   0.369 \\
2019    &   0.261   &   0.225   &   0.263 \\
2020    &   0.172   &   0.260   &   0.311 \\
2021    &   0.191   &   0.279   &   0.363 \\
\midrule
\textbf{Mean} & \textbf{0.227} & \textbf{0.295} & \textbf{0.326} \\
\bottomrule
\end{tabular}
\end{table}

The variation across cutoff years reflects the structural disruptions present in the evaluation periods rather than instability in the model itself. The 2018 cutoff produces the highest RMSE at horizon +2 (0.416), corresponding to projections into 2020: the COVID-19 demand shock year. Observed emissions fell sharply across nearly all sectors and countries simultaneously, a pattern that no trend-extrapolation model trained on pre-pandemic data could anticipate, nor would be expected to. The 2019 and 2020 cutoffs, whose evaluation windows partially overlap with the post-shock recovery period, show more moderate and stable performance, with RMSE values between 0.172 and 0.311 across horizons.

Table \ref{tab:rolling_origin_sector} decomposes RMSE by sector and horizon, averaged across cutoff years. The sectoral ordering is consistent with the uncertainty structure reported in the main text and with the nature of the evaluation periods. Land and Heating \& Cooling show the lowest errors at all horizons (0.121--0.203 and 0.129--0.232 respectively), reflecting the structural regularity of these sectors and their relative insulation from the acute demand shocks that dominated the 2020--2022 period. Mobility and Power show the highest errors (0.253--0.471 and 0.305--0.345 respectively), which is expected: Mobility was among the sectors most severely disrupted by COVID-19 lockdowns, and Power underwent rapid structural change driven by both the renewable transition and the 2022 energy crisis, both of which represent genuine departures from pre-cutoff trends rather than their continuation.

\begin{table}[htbp]
\centering
\caption{Rolling-origin evaluation: RMSE by sector and forecast horizon, 
averaged across training cutoffs $T \in \{2018, 2019, 2020, 2021\}$. 
Metrics are in the z-score normalised emission space used during training. 
Sectors are ordered by mean RMSE across horizons.}
\label{tab:rolling_origin_sector}
\begin{tabular}{lccc}
\toprule
\textbf{Sector} & \textbf{Horizon +1} & \textbf{Horizon +2} 
    & \textbf{Horizon +3} \\
\midrule
Land                & 0.121 & 0.168 & 0.203 \\
Heating \& Cooling  & 0.129 & 0.190 & 0.232 \\
Industry            & 0.164 & 0.226 & 0.267 \\
Mobility            & 0.253 & 0.366 & 0.471 \\
Power               & 0.305 & 0.414 & 0.345 \\
Other               & 0.325 & 0.363 & 0.379 \\
\bottomrule
\end{tabular}
\end{table}

Taken together, these results support the internal validity of the pipeline's short-term projections under stable conditions. The monotonic horizon degradation, the concentration of elevated errors around known structural breaks, and the sectoral ordering of errors indicate that the model is learning economically coherent dynamics rather than overfitting to the training period. Where errors are large, they are large for interpretable reasons: they occur in the sectors and years where real-world dynamics departed most sharply from the kind of trend continuation that an inertial trajectory baseline embeds by construction.

\subsection*{Model sensitivity}
\begin{figure}
    \centering
    \includegraphics[width=\linewidth]{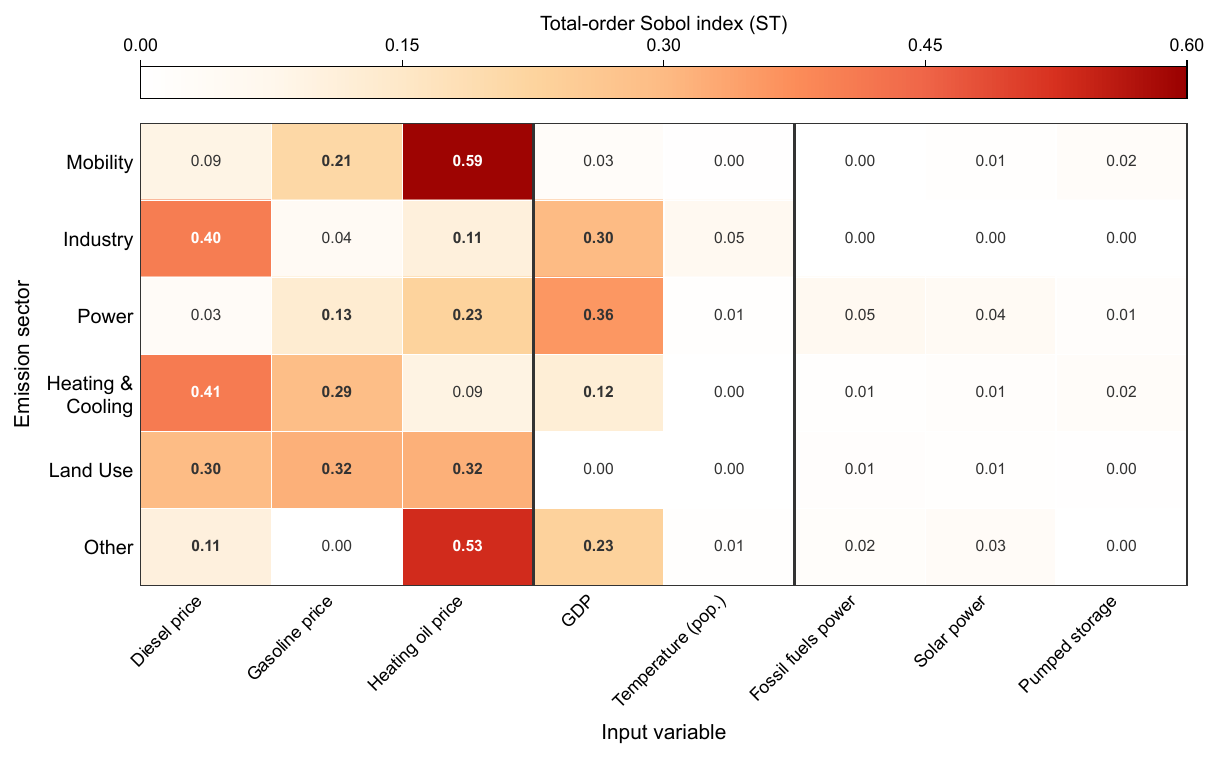}
    \caption{Sensitivity of sectoral CO$_2$ emission predictions to input variables, measured by total-order Sobol indices (ST). Each cell reports the ST index for a given variable-sector pair. Variables are grouped thematically: fuel prices (left), macroeconomic indicators (centre), and electricity generation (right). The analysis is conducted on the full variable set, but only variables appearing in the top five for at least one sector are shown.}
    \label{fig_sobol}
\end{figure}

To characterize the model's internal sensitivity structure and verify that predictions are anchored in economically coherent signals, we conduct a global variance-based sensitivity analysis using total-order Sobol indices \cite{sobol_global_2001, saltelli_variance_2010}, reported in Figure \ref{fig_sobol}. These results describe which inputs most strongly shape the model's output variance; they are not estimates of real-world policy transmission. Given the aggressive dimensionality reduction preceding prediction, we expect the sensitivity structure to be highly concentrated, with most variables carrying negligible indices and predictive variance attributable to a small number of information-dense inputs.

Fuel prices dominate the sensitivity structure across nearly all sectors. This is most naturally interpreted as a consequence of the dimensionality reduction, with fuel prices emerging as the most information-dense observable inputs, co-moving with the business cycle, energy market conditions, and geopolitical shocks simultaneously. This reflects the still strong dependence of national economies on fossil fuels. GDP shows concentrated sensitivity in Industry and Power, consistent with the income elasticity of energy-intensive production. Electricity generation variables carry uniformly low indices, likely because their co-movement with broader economic conditions is already absorbed into the latent representation.

\subsection*{Sectoral emission changes, and model confidence, by country and sector}

\begin{figure}
    \centering
    \adjustbox{width=1.3\linewidth, center}{
        \includegraphics{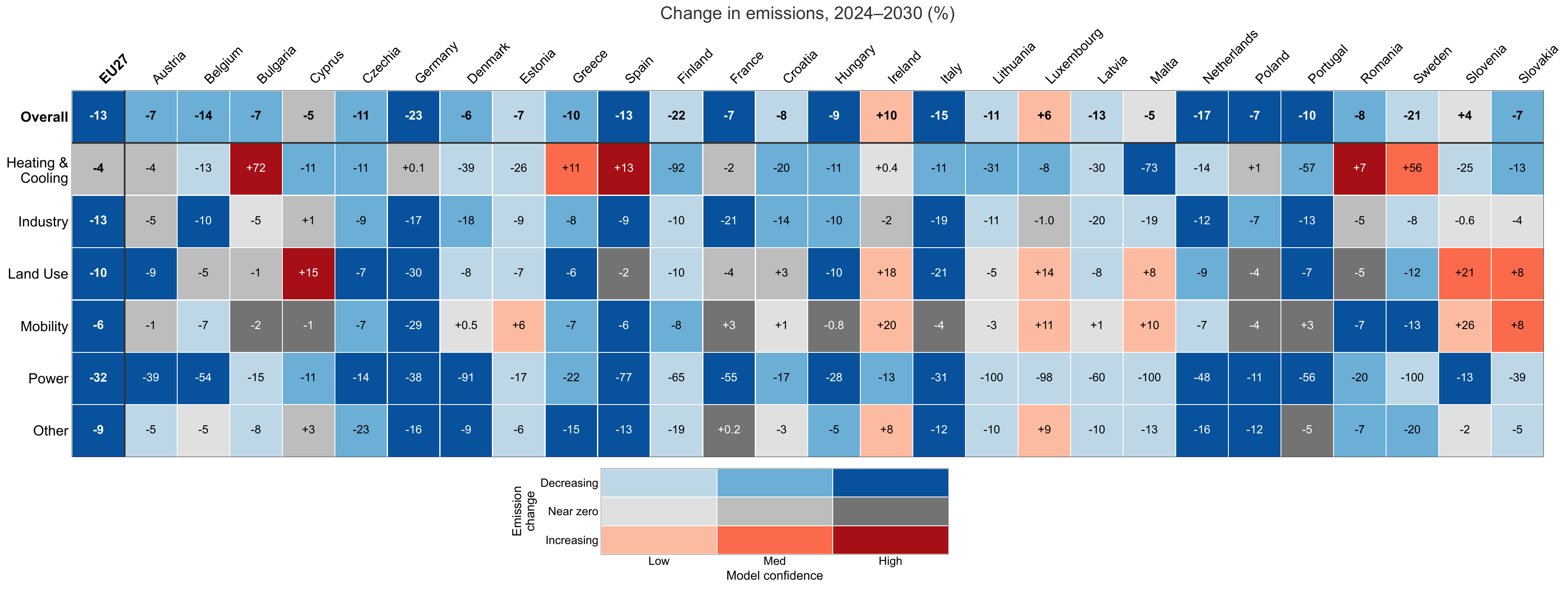}
    }
    \caption{Projected sectoral CO$_2$ emission changes from 2024 to 2030 across EU27 member states. Each cell reports the percentage change in sector-level emissions between 2024 and 2030 under current trends. The bivariate color scheme encodes both the direction of change (blue: decreasing; red: increasing; grey: near zero, defined as within ±5\%) and model confidence (light to dark, derived from the learned uncertainty of the emission predictor). The "Overall" row reports the aggregate change across all sectors. The EU27 column reports the emission-weighted aggregate across all member states. Bold separators distinguish the aggregate row and column from individual entries.}
    \label{fig_sectoral_heatmap}
\end{figure}

Figure \ref{fig_sectoral_heatmap} adds a model confidence dimension to the sectoral patterns described in Section \ref{sec_uneven}. Power carries the highest confidence across member states, reflecting the structural regularity of the renewable transition and the predictability of its near-term trajectory. Mobility confidence is generally medium, including for several countries projecting near-zero changes, indicating genuine uncertainty about whether these tip marginally positive or negative, though the direction of the aggregate trend is robust. Heating \& Cooling and Land Use carry the lowest confidence across most member states. Germany and Italy, identified in the main text as broad-based decarbonizers, maintain this profile at medium-to-high confidence across all six sectors.

\end{document}